\documentclass[pdflatex,sn-mathphys-num]{sn-jnl}

\usepackage{graphicx}%
\usepackage{multirow}%
\usepackage{amsmath,amssymb,amsfonts}%
\usepackage{amsthm}%
\usepackage{mathrsfs}%
\usepackage[title]{appendix}%
\usepackage{xcolor}%
\usepackage{textcomp}%
\usepackage{manyfoot}%
\usepackage{booktabs}%
\usepackage{setspace}
\usepackage{algorithm}%
\usepackage{algorithmicx}%
\usepackage{algpseudocode}%
\usepackage{listings}%
\usepackage{subcaption}%
\newcommand{\para}[1]{\noindent\textbf{#1}\hspace{0.75em}}
\usepackage{soul}

\theoremstyle{thmstyleone}%
\theoremstyle{thmstyletwo}%

\theoremstyle{thmstylethree}%

\begin{document}

\title[Article Title]{SPARS: A Stability-Plasticity-Aware Region Selection Framework for Continual Learning
}


\author*[1]{\fnm{Teif} \sur{Aldaajani}}\email{s44680101@students.tu.edu.sa}

\author*[1]{\fnm{Huda} \sur{Hakami}}\email{hahakami@tu.edu.sa}

\author[1]{\fnm{Maha A.} \sur{Thafar}}\email{m.thafar@tu.edu.sa}


\affil*[1]{
\orgdiv{Department of Computer Science, College of Computers and Information Technology},
\orgname{Taif University},
\orgaddress{
\city{Taif},
\postcode{21944},
\country{Saudi Arabia}
}}


\abstract{Continual learning aims to acquire new knowledge while retaining previously learned knowledge. However, updating shared parameters when learning new tasks often overwrites existing knowledge, causing catastrophic forgetting. Memory replay methods mitigate catastrophic forgetting but rely on previous samples, raising privacy, memory, and scalability concerns. On the other hand, parameter isolation-based methods preserve learned knowledge but may limit knowledge transfer, whereas regularization promotes adaptability but remains susceptible to interference between old and new tasks. To maintain the stability-plasticity trade-off, this paper proposes Stability-Plasticity-Aware Region Selection (SPARS), a replay-free hybrid framework combining global parameter isolation with localized parameter regularization. SPARS estimates parameter importance using either the Fisher Information Matrix (FIM) or a novel Confidence Score (CS) estimator that leverages logit sensitivity and achieves comparable performance to FIM without requiring labelled data. Based on parameter importance, SPARS partitions parameters into three importance-aware regions namely, winning, neighboring, and non-important parameter regions. Each region receives a different adaptation strategy: selective freezing, soft regularization, or unrestricted learning based on its role in the stability-plasticity trade-off. To mitigate classifier bias in class-incremental learning, we employ Quadratic Discriminant Analysis to improve performance on previously learned classes .Experiments on CIFAR-10 and TinyImageNet demonstrate consistent improvement in average incremental accuracy over existing isolation- and regularization-based methods across task- and class-incremental learning. On TinyImageNet, SPARS achieves the best performance with 6\% improvement for class-incremental learning while retaining over 60\% of parameters for future tasks.}




\maketitle

\section{Introduction}\label{sec1}

As the real world is inherently dynamic, data are continuously generated and evolve over time, violating the conventional independent and identically distributed (i.i.d.) assumption on which most deep learning models rely. 
Although deep neural networks (DNNs) have achieved remarkable success across a wide range of applications, they struggle to adapt to continuously evolving data distributions after deployment. 
Continual Learning (CL) aims to enable models to acquire new knowledge sequentially while preserving previously learned knowledge. 
A fundamental challenge in CL is catastrophic forgetting (CF) ~\cite{mccloskey1989catastrophic}, where learning new tasks causes the model to forget previously acquired knowledge. Catastrophic forgetting arises when optimization on newly arriving data modifies parameters that are important for previously learned tasks, thereby disrupting established representations and degrading performance on earlier tasks.

Mitigating catastrophic forgetting is essential for enabling models to adapt to changing data distributions while retaining knowledge acquired from previous tasks.
Existing CL methods can be broadly categorized into memory replay, regularization-based, and parameter-isolation approaches.
Memory replay methods retain a subset of samples from previous tasks and replay them when learning new tasks~\cite{rebuffi2017icarl, titsias2020functional, wang2022memory, yan2022learning}.
Although achieving State-of-the-art performance, these methods, storing previous data introduces privacy and regulatory concerns, particularly in restricted environments, while also increasing memory and computational costs.
Memory-free approaches avoid storing previous samples but face important trade-offs.
Regularization-based methods constrain updates to parameters considered important for previous tasks, but such constraints may be insufficient to preserve learned representations over long task sequences\cite{kirkpatrick2017overcoming, zenke2017continual, aljundi2018memory}. 
In contrast, parameter isolation methods protect previous knowledge by restricting different parameters to different tasks, but progressively consume available model capacity and can eventually lead to model saturation as the number of tasks increases~\cite{mallya2018packnet, mallya2018piggyback, serra2018hard, gurbuz2022nispa}.

 Despite the individual limitations of different CL approaches, their performance is also influenced by the incremental learning setting in which they are evaluated. Incremental learning is generally classified into two settings, namely task-incremental learning (TIL) and class-incremental learning (CIL)\cite{van2022three}.
 Among these settings, CIL is particularly challenging because task identity is unavailable during inference. Therefore, in addition to CF, it suffers from classifier bias, whereby the classifier favors recently learned classes because samples from previous tasks are imbalanced in memory replay methods or unavailable in memory-free methods.
 Some prior work has addressed this issue through model-output calibration \cite{wu2019bic,lin2024class, chen2026unidtr}, while other approaches require additional training procedures or optimization steps \cite{belouadah2019il2m,zhao2019maintaining}. However, these approaches may not be suitable for real-world applications due to either their additional computational cost or the limited reliability of the calibration process.

Building on these limitations, we propose Stability-Plasticity-Aware Region Selection (SPARS), a novel replay-free continual learning framework that explores the underexamined combination of importance-based global parameter isolation and localized parameter regularization. 
Unlike existing approaches that either constrain all important parameters or isolate model components without accounting for their functional importance, SPARS selectively preserves critical regions while allowing less important regions to adapt to new knowledge. 
In particular, we partition the network into three importance-aware regions (winning, neighboring and non-important) and enables each region to adapt according to its role in the stability–plasticity trade-off.
Moreover, to estimate parameter importance in privacy-sensitive environments where access to data is restricted, we propose a novel estimator, termed Confidence Score (CS), which leverages model confidence without requiring labeled data.
To address classifier bias in CIL, we employ Quadratic Discriminant Analysis (QDA) as an inference-time discriminative classifier, eliminating the need for model-output calibration or storing samples from previous tasks. 
The framework is designed as a plug-and-play CL method that can be applied to any pretrained model to support additional tasks. 
It also avoids model saturation, making it more scalable than previous methods.
Extensive experiments are conducted
to demonstrate the effectiveness of the proposed framework against various baselines, along with exploratory studies of its different components.
Comparison experiment results show that our proposed SPARS achieves competitive performance on CIFAR-10 and superior performance on TinyImageNet, with a 6\% improvement over the strongest baselines.
. 

The key contributions of our work are summarized as follows:
\begin{itemize}
 \item Propose SPARS, a replay-free and hybrid CL framework that combines global parameter isolation with local parameter regularization through a region-aware adaptation strategy.

 \item  Introduce a novel Confidence Score (CS)-based parameter importance estimator that leverages logit sensitivity of the model output and provides a label-free alternative to the Fisher Information Matrix (FIM) without requiring access to original training data.
 \item  Mitigate classifier bias in CIL by adopting Quadratic Discriminant Analysis (QDA) as an inference-time classifier, and compare its effectiveness with Linear Discriminant Analysis (LDA) across two benchmark datasets.
 
\end{itemize}

\section{Related Work}
\subsection{Continual Learning Approaches}
Existing continual learning approaches can generally be categorized into
replay-based, parameter regularization-based, parameter isolation-based methods, and hybrid methods. 
Each category addresses catastrophic forgetting from a different perspective, depending on how previously learned knowledge is preserved while adapting to new tasks. 

\medskip
\para {Replay-Based Methods}
Memory replay represents one of the most effective state-of-the-art strategies to mitigate catastrophic forgetting by retaining~\cite{rebuffi2017icarl,lopezpaz2017gradient,chaudhry2019efficient,wang2024uncertainty} or generating past data~\cite{Shin2017DeepGenerativeReplay,Wu2018MemoryReplayGANs,Ostapenko2019LearningToRemember} to rehearse during new task training. 
Early approaches such as Experience Replay~\cite{rolnick2019experience} maintain a memory buffer of previously observed samples, while Incremental Classifier and Representation Learning (iCaRL)~\cite{rebuffi2017icarl} employs exemplar selection and rehearsal together with a distillation objective to preserve knowledge of previously learned classes.
Other methods such as Gradient Episodic Memory (GEM)~\cite{lopezpaz2017gradient,chaudhry2019efficient} store a subset of examples from previously learned tasks and, before updating the model, compare the gradient of the current task with gradients computed from the episodic memory. 
The update is constrained when the gradients conflict, thereby reducing interference with previously acquired knowledge.
Recent replay-based methods, such as Dark Experience Replay (DER)~\cite{Buzzega2020DarkExperienceReplay}, modify traditional experience replay by focusing on soft probabilities (logits) a model produced in the past, thereby providing additional supervision for preserving earlier representations. 
Instead of storing samples from previous tasks, which results in memory overhead, privacy concerns, or data-access restrictions, pseudo-memory replay methods seek to approximate the benefits of rehearsal without maintaining the original training samples. 
Generative replay methods, such as Deep Generative Replay (DGR)~\cite{Shin2017DeepGenerativeReplay} and Memory replay GANs (MeRGAN)~\cite{Wu2018MemoryReplayGANs}, train a generative model to produce synthetic samples resembling earlier tasks, allowing the model to rehearse previous knowledge without directly storing the original examples.

Although replay-based methods effectively reduce forgetting, their dependence on stored or generated samples limits their applicability in scenarios with strict memory or privacy constraints. 
In contrast, parameter-based approaches (regularization or isolation as discussed next) seek to preserve previous knowledge directly within the model parameters, eliminating the need to retain or reconstruct previous data.
This motivates our proposed SPARS framework, which falls into the parameter-based category. 

\medskip

\para{Parameter Regularization-Based Methods} These methods identify parameters that are important for previously learned tasks and penalize changes to those parameters during subsequent training~\cite{kirkpatrick2017overcoming,zenke2017continual,aljundi2018memory, Ebrahimi2019UncertaintyGuidedContinualLearning}. 
Elastic Weight Consolidation (EWC)~\cite{kirkpatrick2017overcoming} estimates parameter importance using the Fisher Information Matrix and imposes stronger penalties on highly important parameters, while Synaptic Intelligence (SI)~\cite{zenke2017continual} estimates parameter importance online based on their contribution to learning throughout training.
Memory Aware Synapses (MAS)~\cite{aljundi2018memory}, in contrast, estimates importance from the sensitivity of the network outputs and can therefore operate without labeled data.
Subsequent approaches have further improved parameter importance estimation, such as RWalk~\cite{chaudhry2018riemannian}, which combines multiple importance measures, and online EWC~\cite{schwarz2018progress}, which maintains a cumulative importance estimate across tasks.
Variational Continual Learning (VCL)~\cite{nguyen2017variational} adopts a Bayesian alternative by using the posterior learned from previous tasks as the prior for the current task, thus constraining the evolution of the model parameters while maintaining uncertainty over their values.

Although these methods can effectively preserve previously acquired knowledge without replay, conventional parameter regularization generally treats parameter importance independently, allowing important parameters to adapt under penalties determined by their estimated importance.

Recent research has therefore explored structured regularization strategies that preserve groups of related parameters to better retain learned representations while maintaining adaptability to new tasks~\cite{jung2020continual,tousside2022group,chen2024pssr}.
However, these approaches typically focus on preserving structured parameter groups rather than explicitly differentiating the degree of adaptation according to the importance of individual parameters and their local relationship to important parameters.
The proposed SPARS addresses this gap by adopting an importance-aware, localized regularization strategy that differentiates the treatment of parameters according to their importance and proximity to important parameters.

\medskip

\para{Parameter Isolation-Based Methods}
This category of CL assumes the independence between tasks so that each task has dedicated parameters~\cite{hurtado2021optimizing, gurbuz2022nispa,jin2023growing}, subnetworks~\cite{Wortsman2020Supermasks,chen2023towards}, masks~\cite{serra2018hard, kang2023forget}, or experts~\cite{zhu2024tame,li2025theory} rather than relying on parameter regularization.  
In general, there is no significant forgetting with this type of method. However, they are not aligned with other desiderata of continual learning, such as knowledge transfer and the trade-off between stability and plasticity. 
A representative approach is Hard Attention to the Task (HAT)~\cite{serra2018hard}, which learns task-specific masks to identify and protect parameters that are important for previously learned tasks.
Building on the idea of allocating different subsets of parameters to different tasks, PackNet~\cite{mallya2018packnet} progressively prunes and freezes parameters after learning each task, thereby reserving the remaining capacity for subsequent tasks.
SupSup~\cite{Wortsman2020Supermasks} takes parameter isolation further by representing each task as a sparse subnetwork within a shared network, allowing multiple tasks to reuse parameters while preventing updates to previously selected subnetworks. 
More recent approaches such as Soft-Winning SubNetworks WSN~\cite{kang2023forget} further explore parameter sharing and task-specific isolation to improve the utilization of the available model capacity.

Despite these advances, isolation-based methods generally impose a hard separation between parameters that are preserved and those available to learn new tasks, which limits their capacity for a long sequence of tasks. 
SPARS builds upon the parameter-isolation paradigm by relaxing the hard separation through applying isolation globally across tasks and regularization locally to important winning parameters.

\medskip
\para{Hybrid  Methods}
Given that excessive emphasis on stability may hinder the acquisition of new knowledge, several continual learning methods integrate mechanisms that preserve previously learned knowledge while simultaneously promoting model plasticity.
These methods combine two or more approaches to address catastrophic forgetting.
Since each category has its own limitations, such combinations can help alleviate the limitations of individual approaches.
Nevertheless, such combinations may introduce new limitations, including additional computational complexity and sensitivity to hyperparameters.
A common approach is to combine regularization-based methods with memory replay to mitigate representation drift while learning new tasks and leveraging stored samples \cite{chaudhry2020using,han2021continual,bhat2022consistency,vahedifar2025,alssum2025forget, manore2026mitigating}.
In addition, regularization has been combined with isolation-based approaches to isolate learned knowledge and integrate it again through knowledge distillation, thereby achieving a balance between acquiring new knowledge and preserving previously learned knowledge \cite{wu2023incremental}.
Isolation-based approaches have also been combined with model-growth approaches, providing a direct means of addressing model saturation.
While isolation can lead to model saturation as the number of tasks increases, model growth helps overcome this limitation and enables the network to scale.
However, model growth introduces additional computational costs, and this combination may limit knowledge sharing across tasks \cite{siddika2026sparse}.
Furthermore, model-growth-based approaches have been combined with memory replay and orthogonal regularization to promote orthogonality between tasks while leveraging memory replay to mitigate bias \cite{ebrahimi2020adversarial,moon2026expandable}.

Despite their potential, these hybrid approaches still inherit some of the limitations of their constituent categories.
Memory replay-based methods require memory buffers, which are limited by available memory capacity and may raise privacy concerns.
Isolation-based approaches can limit knowledge sharing between tasks by keeping them separate.
Regularization-based approaches, meanwhile, typically apply constraints across model parameters regardless of their individual contributions to model stability or plasticity.
Our method aims to address these limitations by handling parameters according to their contribution to model stability and plasticity. Specifically, it incorporates global isolation with local regularization across tasks while avoiding the need for replayed examples, reducing the risk of model saturation, and eliminating the need for model expansion.

\subsection{Bias in Class-Incremental Learning}
\label{sec:CIL_bias}
Class-incremental learning is a challenging CL setting in which task identifiers are unavailable during inference, requiring the model to distinguish among all learned classes without explicit task information.
Consequently, methods that perform well in TIL can suffer substantial performance degradation under CIL, largely due to bias at the representation or classifier level~\cite{zhu2024pass}.
This bias may favor either old or newly introduced classes \cite{shi2024prospective,zhang2020dmc,he2021tale}, with a bias toward new classes being more commonly observed.

Several approaches have been proposed to address the bias issue in CIL, including correction-based, out-of-distribution (OOD)-based, and inference-based methods. 
These approaches differ in where and how bias is addressed: some operate primarily during inference~\cite{lin2024class,rebuffi2017icarl}, whereas others introduce additional training procedures or optimization objectives~\cite{belouadah2019il2m,zhao2019maintaining,wu2019bic}.
Correction-based methods employ explicit bias correction mechanisms, such as lightweight, learnable layers~\cite{wu2019bic} or calibration of model outputs and class weight norms~\cite{zhao2019maintaining,belouadah2019il2m}. 
However, such methods may have limited effectiveness because they primarily compensate for bias at the output level, rather than addressing its underlying causes~\cite{wu2019bic,zhao2019maintaining}.

OOD-based methods~\cite{lin2024class,kim2022theoretical} formulate CIL as two complementary subproblems: within-task prediction, which discriminates among classes belonging to the same task, and task identification, which is treated as an OOD detection problem.
Although this strategy avoids explicit output calibration and its associated limitations, it introduces additional inference complexity and latency because prediction effectively involves two stages.
More importantly, it implicitly attempts to recover task identity at inference time, which is unavailable by definition in the standard CIL setting.

Inference-time solutions can be attractive in practice because they mitigate bias without requiring additional model training or modification of the learned parameters~\cite{mai2021supervised,yang2022continual, momeni2024continual}
One representative approach is presented in~\cite{rebuffi2017icarl}, which replaces the conventional Softmax classifier with a nearest-mean-of-exemplars (NME) classifier. 
However, relying solely on class means may be insufficient for deep neural networks, as their learned feature representations can exhibit complex distributions that are not adequately captured by a single class centroid.
Moreover, in continual learning, the feature distributions of previously learned classes may shift as new classes are introduced, making their original class means less reliable.

Overall, existing bias-mitigation approaches either rely on output-level calibration, require additional statistical information or replay, or introduce inference complexity through implicit task identification.
To address these limitations, our proposed SPARS framework employs a simple inference-time method that replaces the original classifier with a Quadratic Discriminant Analysis classifier during inference. 
Unlike the NME method, QDA models both class means and covariance matrices, enabling it to capture richer class distributions and construct more flexible decision boundaries.
The deep network is therefore used primarily as a feature extractor, while the original classifier is discarded during inference and replaced by QDA. This design provides a practical inference-time solution capable of modeling complex feature distributions without requiring explicit task identification or output-level calibration.

\section{Methodology}
\subsection{Preliminary}
Continual learning is commonly studied under two settings: task-incremental learning and class-incremental learning. 
In both settings, the model learns a sequence of tasks, \(T = [t_1, t_2, t_3, \ldots, t_n]\), where each task \(t\) contains a set of classes, \(C_t\), and their corresponding labels, \(Y_t\). 
Label sets are assumed to be disjoint across tasks.
The model consists of a backbone network and one or more classification heads.
In TIL, the model employs a set of task-specific heads, \(H = [h_1, h_2, h_3, \ldots, h_n]\), where each head \(h_i\) corresponds to a specific task and contains output nodes representing the classes associated with that task.
In CIL, the model employs a single shared classification head that contains all classes learned up to the current task and consequently expands as new classes are introduced.

In TIL, the task identity is available at inference time, enabling the model to select the corresponding task-specific head for prediction.
However, the task identity is often unavailable in real-world applications, making TIL less practical in such scenarios.
In contrast, CIL is task-agnostic, requiring the model to classify each input sample into one of the classes learned across all tasks.
Both TIL and CIL are affected by catastrophic forgetting, where knowledge acquired from previous tasks is degraded as the model learns new tasks.
In addition, CIL is susceptible to classifier bias as discussed in~\autoref{sec:CIL_bias}, whereby the classification layer tends to favor recently learned classes over those encountered in earlier tasks.

\subsection{Stability-Plasticity-Aware Region Selection Framework}
\begin{figure}[htbp]   

 \centering

 \includegraphics[width=0.9\textwidth]{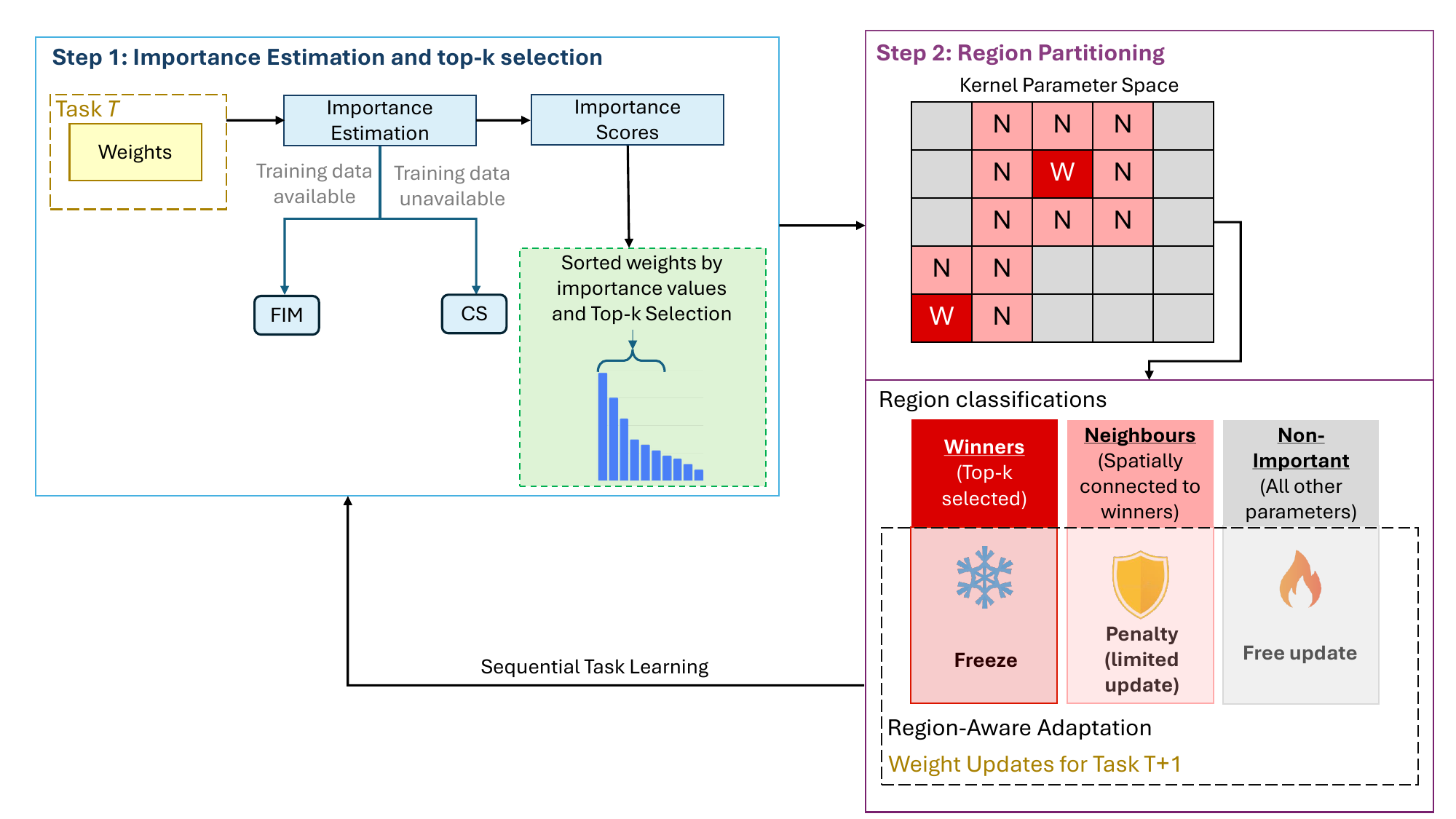}

 \caption{Overview of the SPARS Framework. The framework consists of parameter importance estimation, Top-k parameter selection, partitioning model parameters into winners (W), neighboring (N), and non-important regions, followed a region-aware parameter adaptation strategy.\label{fig:framework}}

\end{figure}

Our proposed framework, SPARS, mitigates CF by combining local parameter regularization with global parameter isolation.
In contrast to existing isolation-based methods that allocate dedicated parameter regions to individual tasks, SPARS performs isolation at the global level by identifying a single shared region containing the most important parameters across all tasks.
This design enables the model to preserve previously acquired shared knowledge more effectively while reducing the risk of network saturation caused by continuously allocating separate parameter regions to new tasks.
Unlike conventional regularization-based methods, which generally impose constraints on a broad set of network parameters, SPARS applies regularization locally to specific parameter regions according to their spatial proximity to important parameters.
This localized regularization provides a soft constraint on these regions while retaining sufficient flexibility for adaptation to new tasks. 
As illustrated in~\autoref{fig:framework}, the proposed framework consists of two main stages: (1) parameter importance estimation (Section \ref{sec:stage1}), which assigns an importance score to each parameter, and (2) importance-based region partitioning (Section \ref{sec:stage2}), which uses these importance scores to divide the network into three importance-aware regions.
The proposed QDA approach for mitigating bias at inference time in the CIL setting is discussed in Section~\ref{sec:mitigate_CIL_bias}. 
Our code is available at \url{https://github.com/teifaldaajani1999-sys/SPARS.git}

\subsubsection{Parameter Importance Estimation}
\label{sec:stage1}
Parameter importance estimation quantifies the sensitivity of a neural network's output to perturbations of its parameters, thereby identifying parameters that are critical for preserving previously acquired knowledge.
Specifically, the importance of a parameter reflects the extent to which a small change in that parameter alters the model's output behavior.
Parameters with high importance scores are considered more critical to the current solution, whereas parameters with low scores are less influential and can therefore be adapted more freely.
This distinction is particularly useful in continual learning, where preserving parameters that encode previously acquired knowledge must be balanced against the need to adapt the model to new tasks.
Moreover, the overparameterization of deep neural networks provides multiple parameter configurations that can achieve comparable performance.
SPARS exploits this flexibility by selectively preserving highly important parameters while allowing less important parameters to adapt to subsequent tasks, thereby improving the stability--plasticity trade-off without requiring access to data from previously learned tasks.

One common approach to estimating parameter importance in CL is the Fisher Information Matrix (FIM) \cite{kirkpatrick2017overcoming}.
From a Bayesian perspective, the model parameters after training encode the knowledge captured by the posterior distribution. 
To estimate how much information each parameter retains, this distribution is analyzed. 
However, because computing the posterior distribution directly is computationally expensive, it is commonly approximated by a Gaussian distribution with mean and variance, where the FIM provides an estimate of the inverse parameter variance. 
This approximation helps to evaluate the importance of each parameter for the task. 
The gradient of the loss function of task $T$, after training, is computed with respect to each parameter. 
A higher magnitude of the gradient indicates greater significance of the parameters, as denoted by the following equation:
    \begin{equation}
     F_i \approx \frac{1}{N} \sum_{n=1}^{N} 
     \left( \frac{\partial \log p(y_n | x_n, \theta)}{\partial \theta_i} \right)^2
    \end{equation}
    
where \(F_i\) denotes the estimated importance of parameter \(\theta_i\), \(N\) is the number of samples, \(x_n\) and \(y_n\) are the \(n\)-th input and its corresponding label, respectively, \(p(y_n \mid x_n, \theta)\) represents the model's predicted probability for the correct class, and \(\frac{\partial \log p(y_n \mid x_n, \theta)}{\partial \theta_i}\) measures the sensitivity of the log-likelihood with respect to \(\theta_i\).

However, empirical FIM-based importance estimation typically requires access to labeled training examples. 
In real-world scenarios, access to the original training set may be infeasible due to privacy concerns, storage limitations, or online learning settings where past data are no longer available. 
This creates a bottleneck for our framework. 
We therefore propose a confidence-based importance score that estimates parameter importance without requiring ground-truth labels. Instead of measuring the sensitivity of the loss with respect to each parameter, this score measures the sensitivity of the model's confidence in its highest probability predicted class to small parameter perturbations. Parameters that strongly affect the model's confidence are considered more important, as they have a larger influence on the model's prediction behavior. 
This formulation is illustrated in the equation below.
\begin{equation}
    CS_i = \frac{1}{N} \sum_{n=1}^{N}
    \left(
    \frac{\partial F(x_n; \theta)}{\partial \theta_{i}}
    \right)^{2}
    \end {equation}
where \(CS_i\) denotes the estimated importance of parameter \(\theta_i\), \(N\) is the number of unseen samples used for evaluation, \(x_n\) is the \(n\)-th input sample, \(F(x_n; \theta)\) represents the logit corresponding to the model's most confident class, and \(\frac{\partial F(x_n; \theta)}{\partial \theta_i}\) measures the sensitivity of this confidence score with respect to \(\theta_i\).

\subsubsection{Importance-Based Region Partitioning}
\label{sec:stage2}
To mitigate catastrophic forgetting while maintaining the stability--plasticity trade-off, the importance scores are utilized to partition the network into three distinct regions. Unlike existing continual learning methods that treat the entire network uniformly, our approach divides the network into three shared regions across tasks: The winning parameter region, the neighboring parameter region, and the non-important parameter region. Each region is then handled differently according to its role in preserving prior knowledge and adapting to new tasks.

\begin{figure}[H]
    \centering
    \includegraphics[width=0.95\linewidth]{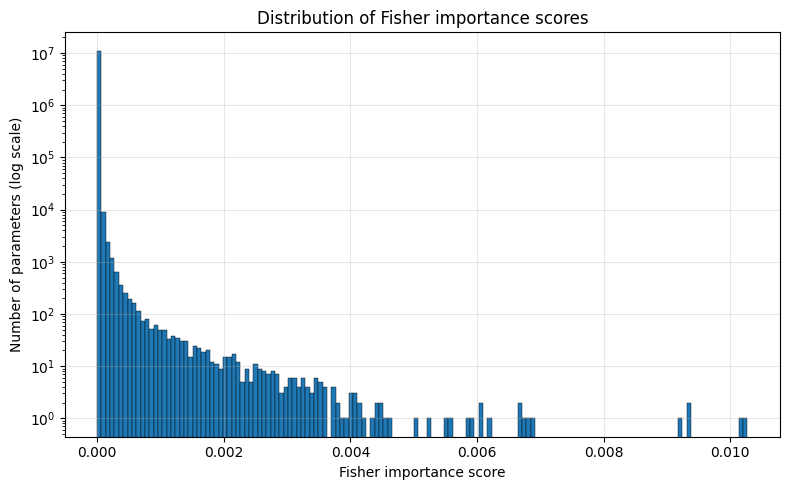}
    \caption{Distribution of Fisher importance scores across all model parameters, showing a long-tailed distribution in which only a small subset of parameters has high importance.}
    \label{full_fisher_dist}
\end{figure}

The motivation behind this design is based on the observed distribution of Fisher values, or importance scores, shown in~\autoref{full_fisher_dist}, which exhibits a long-tailed distribution in which only a small fraction of parameters receives high importance scores. This observation supports the hypothesis that model parameters do not contribute equally to preserving learned knowledge. Accordingly, we define the important region by selecting the top-\(k\) parameters with the highest importance scores across the model. To preserve the spatial structure of the learned representation, we define a neighboring region that consists of parameters adjacent to those in the important region, while all remaining parameters are assigned to the non-important region. Thus, the first region contains the most important parameters from previous tasks, the second region includes parameters spatially adjacent to these important parameters, and the third region consists of parameters considered non-important. 
Let \(I_i\) denote the importance score of parameter \(\theta_i\), and let \(M_i\) denote its update mask coefficient. 
We define the winning-parameter set \(\mathcal{W}_k\) as the set of parameters with the top-\(k\) highest importance scores. 
We also define \(\mathcal{N}(R_1)\) as the set of parameters neighboring the important region \(R_1\). 
The parameters are partitioned into three regions as follows:

\[
\mathcal{R}(\theta_i) =
\left\{
\begin{array}{ll}
R_1, & \theta_i \in \mathcal{W}_k,\\[6pt]
R_2, & \theta_i \in \mathcal{N}(R_1),\\[6pt]
R_3, & \text{otherwise}
\end{array}
\right.
\]

where \(R_1\) represents the important region, \(R_2\) represents the neighboring region, and \(R_3\) represents the non-important region. The update mask coefficient \(M_i\) is then assigned according to the region of each parameter:

\[
M_i =
\left\{
\begin{array}{ll}
0, & \theta_i \in R_1,\\[6pt]
\alpha_i, & \theta_i \in R_2,\\[6pt]
1, & \theta_i \in R_3
\end{array}
\right.
\quad \text{where } 0 < \alpha_i < 1.
\]

A value of \(M_i = 0\) indicates that the parameter is frozen, whereas \(M_i = 1\) indicates that the parameter can be freely updated. Intermediate values, \(0 < \alpha_i < 1\), represent partial update freedom, where the degree of regularization varies according to the parameter importance. 

\medskip
\para{Important Parameter Region (Winning Parameters)}
This region contains the parameters that receive the highest importance scores for previously learned tasks and must remain stable during the training of new tasks. To identify this region, we define top-\(k\) as a hyperparameter that selects the most important weights, i.e., those with the highest importance values, which strongly influence the model's behavior on previous tasks. This region is frozen because it encodes essential knowledge that the model must preserve; therefore, the parameters within this region are not updated during the learning of a new task. This prevents a significant performance drop on previous tasks after learning subsequent tasks. Overall, this region follows the principle of isolation-based methods; however, it operates globally rather than locally. Specifically, the important parameters accumulated across tasks are consolidated into this region and isolated from further updates.

\medskip
\para{Neighboring Parameter Region (Latently Influential Parameters)}
This region contains the neighboring parameters of the winning ones (i.e., the highly important parameters). For each winning parameter $\theta_{o,i,h,w}$, the neighboring parameters are selected from the same convolutional kernel by keeping the output and input channel indices ($o$ and $i$) fixed, while allowing the kernel coordinates ($h$ and $w$) to vary by at most one position, resulting in the eight adjacent parameters. Although these weights may not have high importance scores individually, they are closely connected to and influenced by the winning weights due to their shared input context and joint contribution to the activation of the same neuron. Therefore, regardless of their individual importance values, they still contribute to neuron activation when connected to the winning parameters. This region is subject to a regularization penalty that prevents the parameters from deviating excessively from their previous values, thereby keeping them close to the optimal configuration learned for prior tasks. Moreover, this constraint enables the neighboring parameters to preserve the influence of both the winning parameters and previously learned tasks, while still allowing limited flexibility for adaptation to new tasks. Consequently, these parameters act as a bridge that facilitates knowledge transfer in both backward and forward directions. This region follows the principles of regularization-based approaches, but applies them selectively to specific parameters rather than uniformly across the entire network.

\medskip
\para{Non-Important Parameter Region (Plastic Parameters)}
This region includes parameters with low importance scores, indicating that changes to these parameters have minimal impact on previously learned tasks. Consequently, this region consists of all remaining parameters and provides high plasticity, while the previous regions contribute to maintaining stability. These parameters can be updated freely during learning, allowing the model to adapt to new tasks without compromising the stability preserved by the first two regions.

The three regions described above collectively define the proposed partitioning strategy.
Algorithm \autoref{alg:region-partition}  summarizes the complete framework, including the region partitioning stage after learning task $t_i$ and the region-aware adaptation (fine-tuning) stage for task $t_i+1$.

\medskip
\para{Region Accumulation Across Tasks}
The regions are updated cumulatively across tasks. Specifically, the winner region is defined as the union of winner parameters identified from all previously learned tasks and remains frozen during training on the current task. Likewise, the neighboring region is constructed as the union of neighboring parameters identified across previous tasks. If a parameter is identified as a neighbor in multiple tasks, it is stored only once, and its associated weight is set to the maximum value observed across those tasks. Parameters are assigned exclusively to either the winning region or the neighboring region. Therefore, if a parameter belongs to both sets, it is assigned to the winner region. This design maintains a clear separation between parameter regions throughout incremental learning.

\begin{algorithm}[t]
\small
\caption{\textbf{Importance-Based Region Partitioning and  Region-Aware Adaptation}
\footnotesize }
\label{alg:region-partition}

\begin{spacing}{1.15}
\begin{algorithmic}[1]

\Require Trained model after task {$t_i \in T$} with parameters $\theta$; 
$K$: Top-K hyperparameter; regularization weight $\lambda$.

\vspace{0.5em}

\State \textbf{Phase 1: Region Partitioning After Task $t_i$}

\ForAll{$p \in \theta$}
    \State Compute $F(p)$ 
    \hfill // The importance function
\EndFor

\State $L \gets$ Sorted parameters in descending order of $F(p)$

\State $W \gets L[1..K]$
\hfill // Winning Parameters

\State $N \gets \emptyset$
\hfill // Neighboring Parameters

\ForAll{$w_{ij} \in W$}

    \If{$w_{ij}$ in a convolutional layer}

        \State $N \gets N \cup 
       \{w_{i\pm1,j}, w_{i,j\pm1}, w_{i\pm1,j\pm1}\}$ \hfill // Set of eight spatially adjacent parameters. 

    \EndIf

\EndFor

\vspace{0.5em}

\State \textbf{Phase 2:  Region-Aware Adaptation}

\ForAll{$p \in \theta$}

    \If{$p \in W$}

        \State $p.\text{grad} \gets 0$
        \hfill // Freeze winning parameters

    \ElsIf{$p \in N$}

        \State $p.\text{grad} \gets \text{grad} + (\lambda \cdot F(p) \cdot (p_{\text{old}} - p_{\text{new}})^2$)
        \hfill{// Soft regularization}

    \Else

        \State $p.\text{grad} \gets \text{grad}$
        \hfill // Full plasticity

    \EndIf

\EndFor

\end{algorithmic}
\end{spacing}
\end{algorithm}

\subsubsection{Mitigating Classifier Bias in CIL}
\label{sec:mitigate_CIL_bias}
Classifier bias is a well-known issue in class-incremental learning, where the model tends to favor recently learned classes over previously learned ones. This bias primarily arises from the imbalance between old and new class samples in replay-based methods. In memory-free methods, where samples from previous tasks are unavailable, the absence of old data makes the problem even more challenging.

To alleviate classifier bias within the proposed memory-free framework, we replace the conventional fully connected classifier during inference with a discriminant classifier based on Mahalanobis distance. Specifically, the backbone network is used solely as a feature extractor to produce discriminative feature representations, while the final FC classification layer is discarded. Classification is then performed using the extracted features together with the estimated class means and covariance matrices. 
We investigate both LDA and QDA as the discriminant classifiers \cite{hastie2009elements}. 
LDA assumes a shared covariance matrix across classes, whereas QDA estimates an individual covariance matrix for each class, allowing more flexible decision boundaries. 
The LDA discriminant function is defined as follows:

\begin{equation}
\label{eq1}
\delta_k(x)
=
x^T \Sigma^{-1} \mu_k
-
\frac{1}{2}\mu_k^T \Sigma^{-1} \mu_k
+
\log P(y = k)
\end{equation}
where $x$ denotes the input feature vector, $\mu_k$ is the mean feature vector of class $k$, $\Sigma$ is the pooled covariance matrix shared across all classes, and $P(y = k)$ represents the prior probability of class $k$. Unlike LDA, QDA estimates a separate covariance matrix for each class, leading to the following discriminant function.

\begin{equation}
\label{eq2}
\delta_k(x) =
-\frac{1}{2}\log |\Sigma_k|
-\frac{1}{2}(x-\mu_k)^T \Sigma_k^{-1}(x-\mu_k)
+\log \pi_k
\end{equation}

where $x$ denotes the input feature vector, $\mu_k$ is the mean feature vector of class $k$, $\Sigma_k$ is the covariance matrix of class $k$, $|\Sigma_k|$ denotes the determinant of the covariance matrix $\Sigma_k$, and $\pi_k$ represents the prior probability of class $k$.
The comparative performance of LDA and QDA is evaluated in section \ref{CIL Discriminant Classifier Analysis}

\section{Experiments }
\label{ES}
\subsection{Experimental Setup}
\para{Datasets}
Following recent work on continual learning, we use two widely adopted datasets, CIFAR-10~\cite{krizhevsky2009learning} and TinyImageNet~\cite{le2015tiny}, for both CIL and TIL settings. 
We use the standard protocol, in which each dataset is partitioned into multiple tasks, each containing a distinct, non-overlapping subset of classes. \autoref{tbl1} summarizes the dataset statistics and the specific splits used in our experiments.
CIFAR-10 consists of 60,000 RGB images of size 32×32 pixels, split into 50,000 training samples and 10,000 test samples. 
Each of the 10 classes contains 5,000 training images and 1,000 test images, with no overlap between classes. 
Following the standard continual learning setting, CIFAR-10 is divided into five tasks, each containing two classes, for both TIL and CIL scenarios~\citep{krizhevsky2009learning}.
TinyImageNet is a subset of the larger ImageNet Large Scale Visual Recognition Challenge (ILSVRC) dataset ~\cite{le2015tiny}. It includes 200 classes, each with 500 training samples and 100 validation/test samples (50 each). 
Each image is 64×64 pixels. It is typically divided into 10 tasks, each containing 20 classes for use in both TIL and CIL scenarios.

\begin{table}[htbp]
\centering
\caption{Dataset statistics and continual learning task splits used in our experiments.}
\label{tbl1}

\begin{tabular}{lccc}
\toprule
\textbf{Dataset} & \textbf{Train} & \textbf{Test} & \textbf{TIL and CIL Splits} \\
\midrule
CIFAR-10 & 50,000 & 10,000 & 5 tasks, each with 2 classes \\
TinyImageNet & 100,000 & 10,000 & 10 tasks, each with 20 classes \\
\bottomrule
\end{tabular}

\end{table}

\para{Evaluation Metrics}
The performance is evaluated using two standard metrics that capture complementary aspects of CL namely, Average Incremental Accuracy (AIA) and Average Forgetting (AF).
AIA measures the model's average classification accuracy across all incremental learning steps. 
At each step, accuracy is computed over all tasks encountered up to that point, and AIA is obtained by averaging these accuracies across all steps.
~\autoref{eq:AIA} defines the AIA metric. 

\begin{equation}
\label{eq:AIA}
        \mathrm{AIA} = \frac{1}{T} \sum_{i=1}^{T} AA_{i}
     \end{equation}    
where $T$ denotes the total
number of tasks, and $AA_i$ denotes the average accuracy on task $i$.

Average forgetting is a direct measure of catastrophic forgetting that measures how much performance on previously learned tasks decreases after the model learns subsequent tasks.
For each task $i$, the forgetting at the final step is defined as the difference between its maximum previously achieved accuracy, $\max_{t \in \{i,\ldots,T-1\}}
 a_{t,i}$, and its accuracy after the final task $a_{T,i}$ as follows:
\begin{equation}
    f_i = \max_{t \in \{i,\ldots,T-1\}}
 a_{t,i} - a_{T,i} 
\end{equation}
The AF is defined across all previously learned tasks as in~\autoref{eq:AF}, where the last task is excluded because there is no subsequent learning step after it from which forgetting can be measured.

\begin{equation}
\label{eq:AF}
    AF = \frac{1}{T-1} \sum_{i=1}^{T-1} f_i
\end{equation}

\medskip

\para{Implementation Details}
Following previous work~\cite{ Buzzega2020DarkExperienceReplay,pourkeshavarzi2022looking,wang2022memory, li2024towards}, we use ResNet18 \cite{he2016deep} as the backbone architecture for all experiments, and it is trained from scratch without pre-training. 
For TIL, we adopt a multi-head setup with five heads for CIFAR-10 and ten heads for TinyImageNet, where each head corresponds to a specific task. For CIL, we use a single-head classifier that starts with the initial number of classes and expands incrementally as new tasks are introduced. 
Following \cite{Buzzega2020DarkExperienceReplay}, we apply standard data augmentation techniques, including random cropping and random horizontal flipping, to the training data for both CIFAR-10 and TinyImageNet. 
The framework is trained using the cross-entropy loss function and optimized with SGD optimization. 
The learning rate is set to $10^{-5}$
for the backbone, while the classifier head is trained with a learning rate of 0.9 to allow faster adaptation to newly introduced tasks. 
The weight decay is set to $5 \times 10^{-9}$
, with a batch size of 32 in all experiments. 
For the CL configuration, the Fisher regularization coefficient is fixed at $\lambda = 2$. 
The top-k value is set to 200K in the TIL setting. 
In the CIL setting, the first task uses 900K, while 300K is used for all subsequent tasks. Experiments are conducted on a single NVIDIA A100 GPU.

\medskip

\para{Baselines}
To assess the effectiveness of SPARS, we compare SPARS against different categories of continual learning methods: regularization-based methods (oEWC~\cite{schwarz2018progress}, SI~\cite{zenke2017continual}, and MAS~\cite{aljundi2018memory}), isolation-based methods (SupSup~\cite{Wortsman2020Supermasks}, WSN~\cite{kang2023forget}, SPG~\cite{konishi2023parameter}, and IBM~\cite{chen2023towards}), and hybrid methods (EDD~\cite{moon2026expandable} and HAL~\cite{chaudhry2020using}). 
Even though our framework is data-free, we also compare it with memory replay-based methods (iCaRL~\cite{rebuffi2017icarl}, DER~\cite{Buzzega2020DarkExperienceReplay}, BiC~\cite{wu2019bic}, WA~\cite{zhao2019maintaining}, TPCIL~\cite{Tao2020TPCIL}, Mnemonics~\cite{liu2020mnemonics}, MER~\cite{riemer2018learning}, GEM~\cite{lopezpaz2017gradient}, and a-GEM~\cite{chaudhry2019efficient}, ) and pseudo-replay methods (DGM~\cite{Ostapenko2019LearningToRemember}, DGR~\cite{Shin2017DeepGenerativeReplay}, and MeRGAN~\cite{Wu2018MemoryReplayGANs}). 
Among the replay-based methods, iCaRL, BiC, and WA are baselines that address the bias issue in the CIL setting.
iCaRL relies on the nearest mean of exemplars to classify samples, BiC adopts a lightweight layer to correct the bias through learnable parameters, and WA aligns the weights of the classifier by aligning the norms of the weight vectors of new classes with those of old classes. We also compare SPARS against an upper-bound JOINT baseline obtained by jointly training on all tasks simultaneously. 
Brief descriptions of these baselines are provided in \autoref{app:baselines}.

\subsection{ Main Results and Analysis} 
\autoref{tab:TIL_CIL_AIA} presents the average incremental accuracy of the proposed method and the baseline methods under both TIL and CIL settings. 
Replay-based methods achieve the highest performance on CIFAR-10 and rank second on TinyImageNet. However, their effectiveness is constrained by the fixed memory-buffer capacity, which limits the number and diversity of stored samples as the number of classes increases. 
Although various memory management strategies can reduce this limitation, replay-based methods remain difficult to scale to long task sequences. Under these conditions, pseudo-replay methods can perform comparatively well by generating samples from previously learned classes. 
In contrast, regularization-based methods show lower performance on both benchmarks, falling behind the other continual learning categories by a considerable margin. This performance gap is more pronounced as the number of tasks and classes increases, because estimating and preserving parameter importance becomes increasingly difficult. 
The proposed SPARS method achieves AIA values of \(89.35\%\) and \(61.61\%\) on CIFAR-10 and TinyImageNet, respectively, under TIL. On CIFAR-10, SPARS remains competitive with the leading replay-based methods despite requiring no stored samples from previous tasks. 
On TinyImageNet, SPARS outperforms all compared categories and achieves approximately \(76\%\) of the JOINT upper bound, demonstrating strong performance under class-scarce conditions. Furthermore, SPARS  outperforms pseudo-replay methods without relying on synthetic data generation, thereby avoiding potential noise and visual artifacts that associated with generated samples and degrade representation quality during incremental learning.

Under the CIL setting, all methods exhibit a noticeable performance decline when moving from CIFAR-10 to TinyImageNet, reflecting the greater difficulty of the latter benchmark. TinyImageNet contains \(20\) times more classes than CIFAR-10, making the classification problem substantially more diverse. Moreover, it provides only \(500\) training examples per class, compared with \(5{,}000\) examples per class in CIFAR-10. Consequently, the performance differences among continual learning categories become smaller on TinyImageNet. Among the compared methods, DGM achieves the highest performance on CIFAR-10,while EDD achieves the second best performance on TinyImageNet. The DGM performance can be attributed to its pseudo-replay mechanism, which helps preserve knowledge of previously learned classes through generated samples. Nevertheless, SPARS achieves \(59.28\%\) and \(36.13\%\) AIA on CIFAR-10 and TinyImageNet, respectively, and surpasses DGM on TinyImageNet by a substantial margin without relying on synthetic replay samples.

Overall, SPARS demonstrates strong performance under both TIL and CIL on TinyImageNet, indicating that it can acquire new knowledge while retaining a substantial portion of previously learned knowledge. It also maintains a favorable stability--plasticity trade-off under long task sequences and challenging benchmark conditions. These results further indicate that methods performing well on CIFAR-10 do not necessarily maintain the same performance on TinyImageNet, where the larger number of classes, the smaller number of samples per class, and the longer task sequence may contribute to a more challenging continual learning setting. Therefore, SPARS appears particularly suitable for continual learning scenarios in which data are limited, and tasks arrive sequentially over time.

\begin{table*}[t]
\caption{Results of the proposed method compared with different baselines in terms of AIA. Bold values indicate the best performance, while “-” denotes experiments that were not reported in the reference papers. All baseline results are reported from~\cite{pourkeshavarzi2022looking}, while for the baselines marked with “*”, their results are taken from the cited papers indicated alongside them.}
\centering
\small
\makebox[\textwidth][c]{%
\begin{tabular}{llcccc}
\hline
\multirow{2}{*}{Category} &
\multirow{2}{*}{Method} &
\multicolumn{2}{c}{TIL} &
\multicolumn{2}{c}{CIL} \\
\cmidrule(lr){3-6}
& & CIFAR-10 & TinyImageNet & CIFAR-10 & TinyImageNet \\
\hline

JOINT & Upper bound & 98.45 & 82.07 & 92.02 & 59.45 \\
\hline

\multirow{4}{*}{Regularization based}
& oEWC & 68.29 & 19.20 & 19.49 & 7.58 \\
& SI  & 68.05 & 36.32 & 19.48 & 6.58 \\
& MAS*~\cite{carta2023projected} & - & 45.08 & - & 27.79 \\
\hline

\multirow{3}{*}{Isolation based}
& Supsup*~\cite{chen2023towards} & - & 40.27 & - & -  \\
& SPG*~\cite{chen2023towards} & - & 21.14 & - & -  \\
& WSN*~\cite{chen2023towards} & - & 49.36 & - & -  \\
& IBM*~\cite{chen2023towards} & - & 52.38 & - & -  \\
\hline
\multirow{3}{*}{Memory replay}
& MER & 93.61 & 48.64 & 57.74 & 9.99 \\
& GEM & 92.16 & -- & 26.20 & -- \\
& a-GEM & 89.48 & 25.33& 22.67 & 8.06 \\
& DER & 93.40 & 51.78 & 70.51 & 17.75 \\
& TPCIL & \textbf{96.67} & 48.41 & 58.72 & 25.03 \\
& iCaRL & 88.22 & 31.55 & 47.55 & 9.38 \\
& BiC & 89.91 & 32.94 & 49.18 & 11.09 \\
& WA & 94.21 & 37.12 & 53.76 & 15.42 \\
& UER*~\cite{wang2024uncertainty} & 93.89 & 44.81 & 68.87 & 13.61 \\
\hline

\multirow{3}{*}{Pseudo memory replay}
& DGM & 89.91 & 61.32 & \textbf{71.94} & 28.45 \\
& DGR & 79.86 & 38.41 & 49.69 & 17.38 \\
& MeRGAN & 84.76 & 58.32 & 66.76 & 23.86 \\
\hline

Hybrid approach
& EDD*~\cite{moon2026expandable} & - & - & 55.13 & 30.11 \\
& HAL*~\cite{moon2026expandable} & - & - & 19.99 & 8.89 \\
& SPARS-FIM (Our) & 89.35 & \textbf{61.61}
& 59.28 & \textbf{36.13} \\
& SPARS-CS (Our) & 89.70 & 60.29
& 59.37 & 36.12 \\
\hline

\end{tabular}}
\label{tab:TIL_CIL_AIA}
\end{table*}

\subsubsection{Effect of the Confidence-Based Importance Score}
We evaluate the effectiveness of the proposed CS importance estimation method on CIFAR-10 and TinyImageNet under both TIL and CIL settings.

As shown in~\autoref{tab:TIL_CIL_AIA}, SPARS-CS achieves performance competitive with SPARS-FIM on both CIFAR-10 and TinyImageNet under TIL and CIL, despite not requiring access to training data or class labels for importance estimation. 
These results indicate that effective parameter importance estimation can be achieved without labelled data, thereby improving the robustness and applicability of SPARS in settings where data access is limited. 
Moreover, the Confidence Score is not specific to SPARS and could potentially be integrated into other continual learning methods that rely on parameter importance estimation.

\subsubsection {CIL Discriminant Classifier Analysis}
\label{CIL Discriminant Classifier Analysis}
\begin{figure}[t]
    \centering

    \begin{subfigure}[b]{0.7\textwidth}
        \centering
        \includegraphics[width=\textwidth]{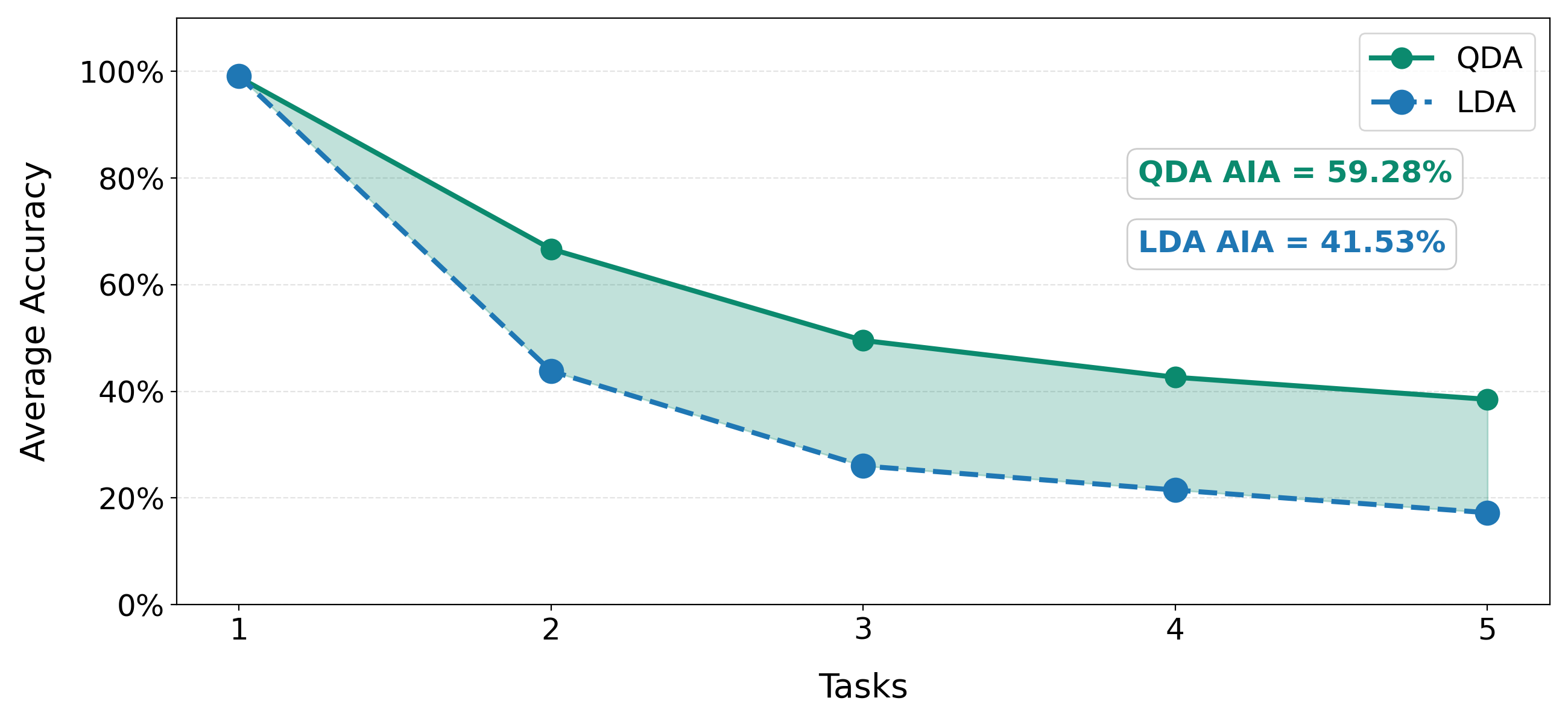}
        \caption{CIFAR-10}
        \label{fig:first}
    \end{subfigure}
    \hfill
    \begin{subfigure}[b]{0.7\textwidth}
        \centering
        \includegraphics[width=\textwidth]{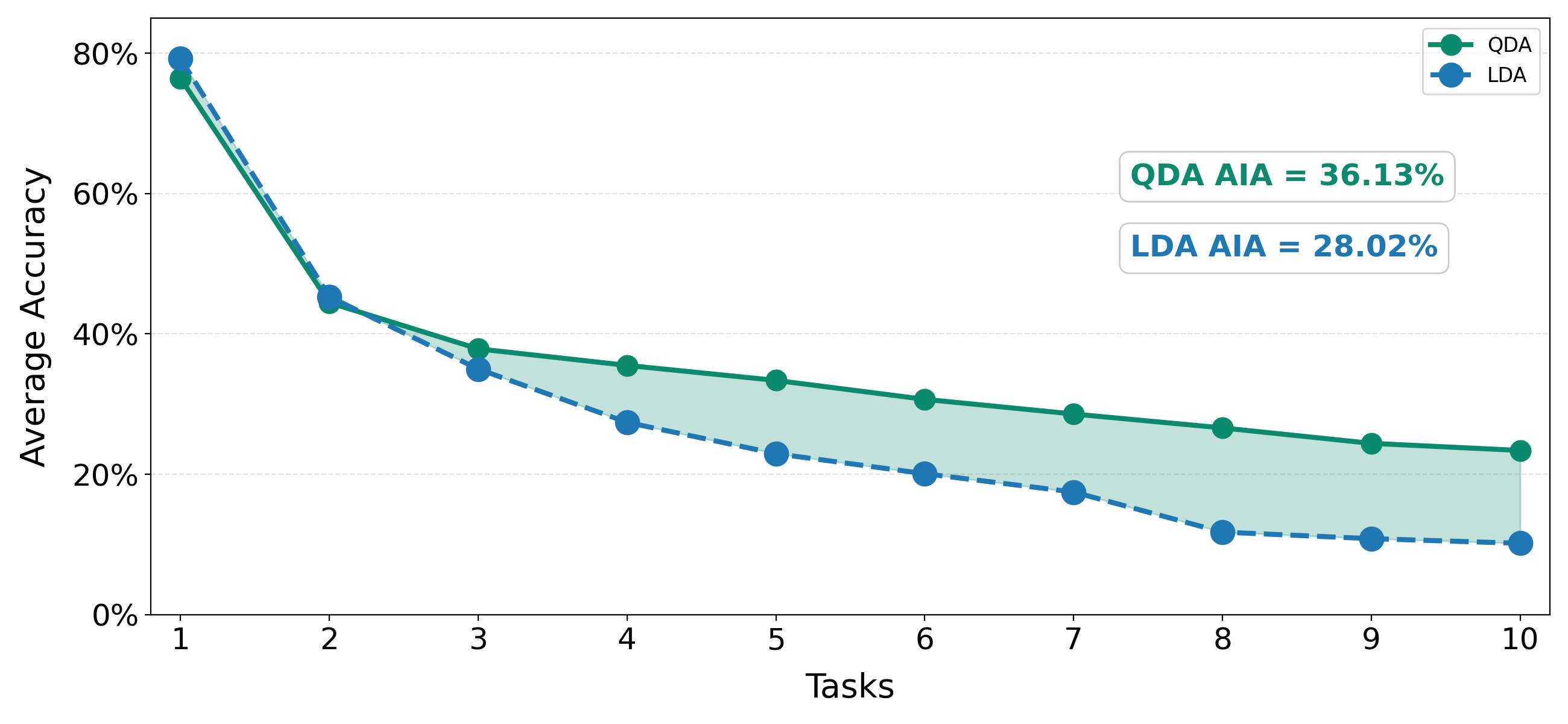}
        \caption{TinyImageNet}
        \label{fig:second}
    \end{subfigure}

    \caption{Per-task Average Accuracy of QDA and LDA across the tasks for CIL on (a) CIFAR-10 and (b) TinyImageNet.}
    \label{fig:comparison_qda_lda}
\end{figure}

\begin{figure}[H]
    \centering
    \includegraphics[width=0.95\linewidth]{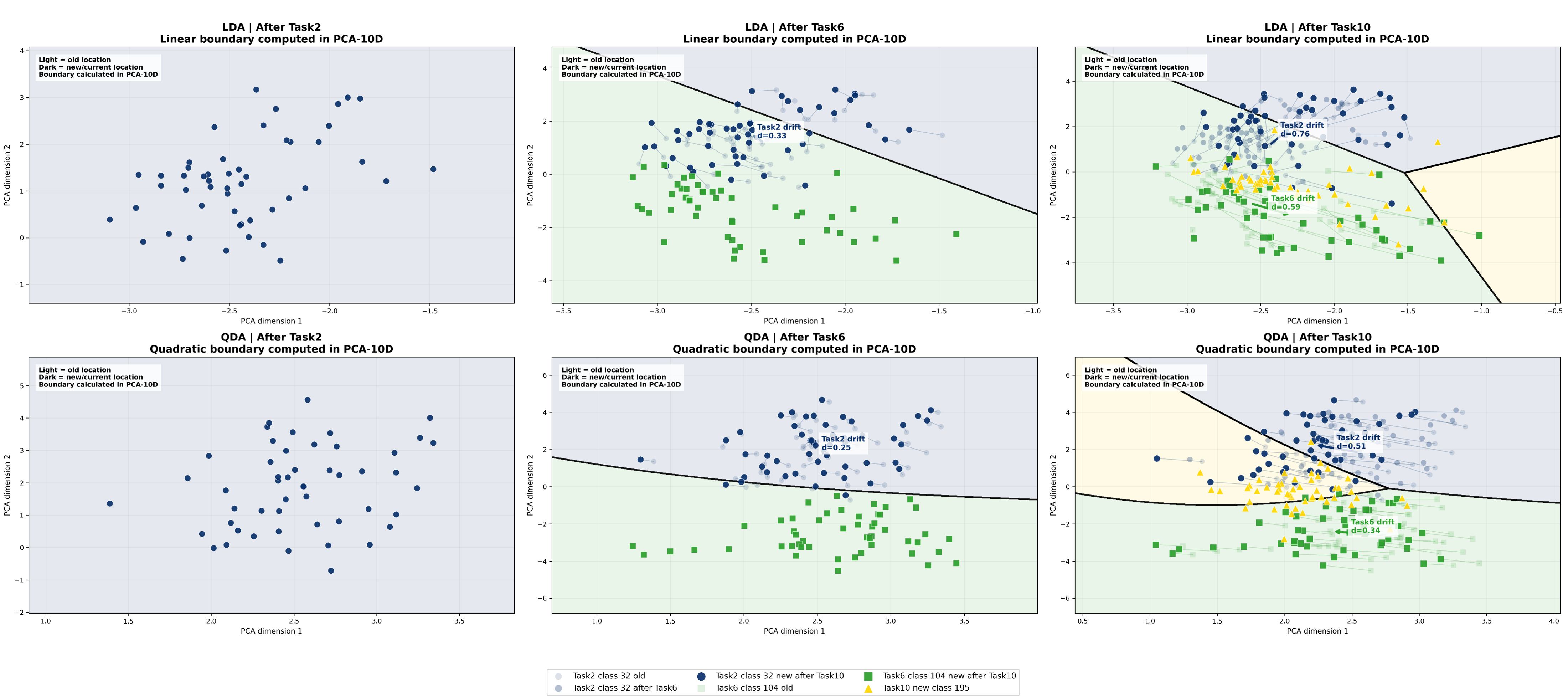}
    \caption{PCA visualization of feature embeddings extracted from the backbone network after tasks 2, 6, and 10 under the CIL setting. Light-colored samples denote earlier feature locations, while darker samples represent their updated positions after subsequent tasks. Arrows visualize the direction and magnitude of feature drift over time.}
    \label{fig: QDA_LDA_visu.png}
\end{figure}

To investigate the bias observed in the CIL setting, alternative inference-time classification methods are evaluated using feature representations extracted from the trained backbone network. Unlike the conventional classifier head, these classifiers utilize the network solely as a feature extractor. Consequently, they provide an opportunity to address bias without requiring norm or logit calibration and without modifying the training dynamics. 
Therefore, Mahalanobis distance-based classifiers, namely LDA and QDA, are investigated in this section as inference-time alternatives. Their performance is first evaluated on CIFAR-10 and TinyImageNet, followed by an analysis of their strengths and limitations in continual learning scenarios. 
\autoref{fig:comparison_qda_lda} shows the AIA results on CIFAR-10 for LDA and QDA. Both LDA and QDA begin with similar AA on the first task. However, starting from the second task, a gap of approximately 30 percentage points appears in favor of QDA, and this gap remains consistent across the subsequent tasks. Overall, the AIA of QDA is higher than that of LDA by 17.75 percentage points. On TinyImageNet, QDA and LDA achieve similar AA during the first two tasks. Afterward, QDA consistently outperforms LDA by approximately 8 percentage points. In addition, the results show greater stability for QDA on both benchmarks. This is because QDA can create more flexible decision boundaries by considering individual covariance matrices for each class, which helps when feature distributions drift over time. Since these boundaries are curved rather than rigid, unlike LDA, they remain more adaptable to new feature space locations. Especially at Task 10, the nonlinear overlap between classes makes LDA struggle to separate them effectively. \autoref{fig: QDA_LDA_visu.png} illustrates the feature space behavior of LDA and QDA under the CIL setting after different sequential tasks. The visualizations were generated using PCA-projected feature representations, with each color representing a different class and the shaded regions indicating the corresponding decision boundaries. The figure shows the decision boundaries of LDA and QDA for three classes: the first class from Task 2, and the second and third classes from Tasks 6 and 10, respectively. The arrows indicate feature drift between earlier and later task states. While noticeable drift occurs for previously learned classes, the class clusters remain partially separable in the learned feature space. LDA fails to capture this evolving geometry, whereas QDA adapts more effectively to nonlinear distributional changes.

Overall, these observations, with the amount of feature drift, suggest that the issue in CIL is not solely caused by feature drift, as the learned representations retain partial robustness. Instead, the results indicate that a significant portion of the performance degradation originates from the classifier method itself, which is consistent with the behavior shown in the figures. The improved separation obtained with QDA suggests that part of the degradation in CIL arises from limitations of the classifier head rather than from complete degradation of the learned representations.

\subsection{Ablation Studies}

\medskip

\begin{figure}[H]
    \centering
\includegraphics[width=0.80\linewidth]{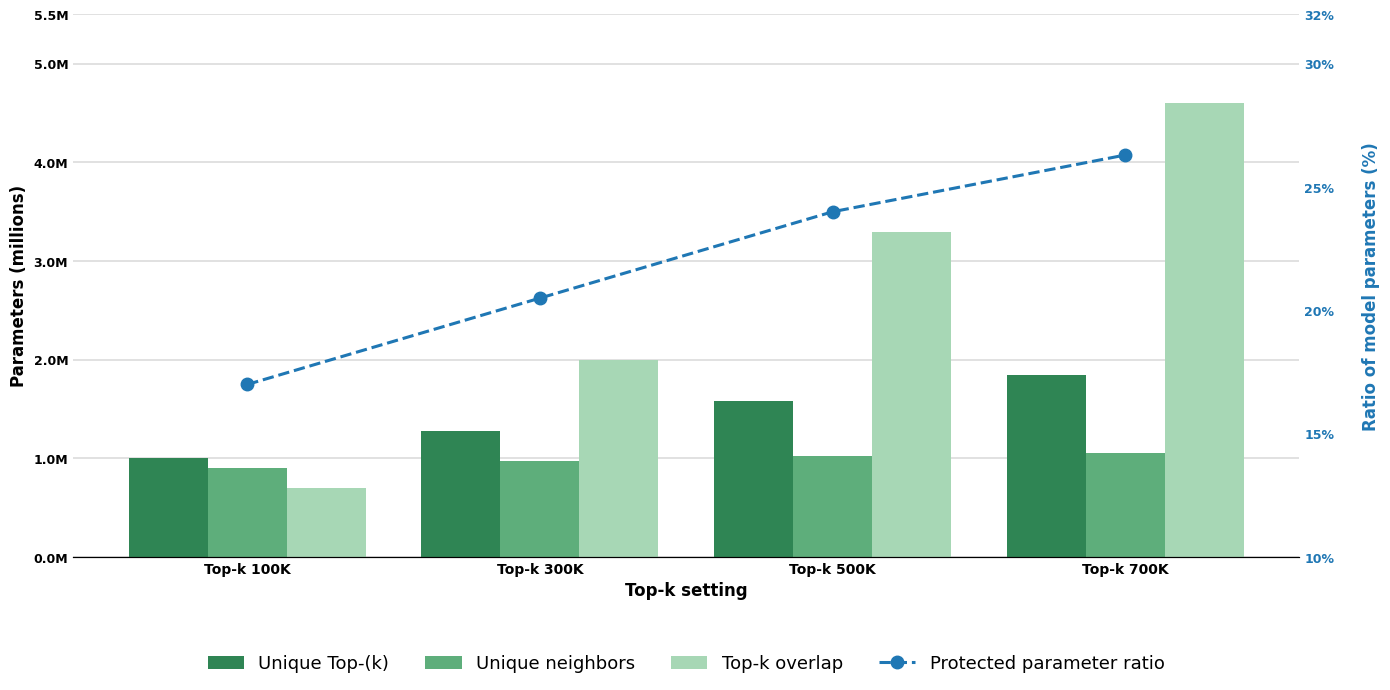}
    \caption{Comparison between the number of unique top-(k) parameters across 10 tasks on TinyImageNet under CIL setting, the number of unique neighboring parameters accumulated across tasks, and the overlap among the top-(k) parameters identified across tasks. The blue curve shows the ratio of the combined top-(k) and neighboring parameters to the total number of model parameters.}
    \label{fig:Top-k}
\end{figure}

\para{Effect of Top-k Values}
To investigate the effect of the top-k parameters, we conducted extensive experiments using the following top-k values: [100k, 300k, 500k, 700k]. Note that the top-k value for the first task remains fixed at 900k, as used in the main results. Preserving this setting allows for a fair comparison among the different top-k values. 
~\autoref{tbl:topk} shows that the 500K configuration achieves the highest AIA among all evaluated configurations. It also produces the lowest AF, indicating improved resistance to catastrophic forgetting. Overall, these results suggest a favorable trade-off between stability and plasticity. Although 500k achieves the best overall performance, the variation in AIA across the evaluated top-k values remains relatively small. This suggests that the proposed framework is generally robust to moderate changes in the top-k setting while maintaining a favorable stability–plasticity trade-off. As shown in \autoref{fig:Top-k}, increasing the top-k value also increases the overlap of top-k parameters across tasks. Therefore, the number of unique top-k parameters remains lower than the total accumulated top-k parameters across all tasks, indicating the existence of shared winner (important) parameters among tasks. This observation directly suggests that these frozen parameters contribute to performance across task sequences, indicating a degree of forward knowledge transfer. In addition, only a subspace of the model parameters is preserved through freezing the top-k (winner) parameters and regularizing neighboring parameters. Furthermore, the percentage of this protected parameter subspace does not exceed 27\% of the total model parameters. As a result, a large portion of the network remains available for adaptation, indicating that the model still retains sufficient capacity to learn additional tasks without saturation.

\begin{table*}
\caption{AIA and AF across different top-$k$ sizes for CIL on TinyImageNet. Higher AIA and lower AF indicate better performance across all 10 tasks. The best result is achieved at top-$k = 500k$.}\label{tbl:topk}
\centering
\small
\begin{tabular}{lllll} 
\toprule
\textbf{Top-k} & \textbf{100K } & \textbf{300K} & \textbf{500K} & \textbf{700K} \\
\midrule
AIA & 36.02  &36.13  & \textbf{36.14} &36.11 \\
AF & 15.86  &15.48  & \textbf{14.60} &14.77 \\
\bottomrule
\end{tabular}
\end{table*}

\medskip

\begin{figure}[t]
    \centering
\includegraphics[width=0.80\linewidth]{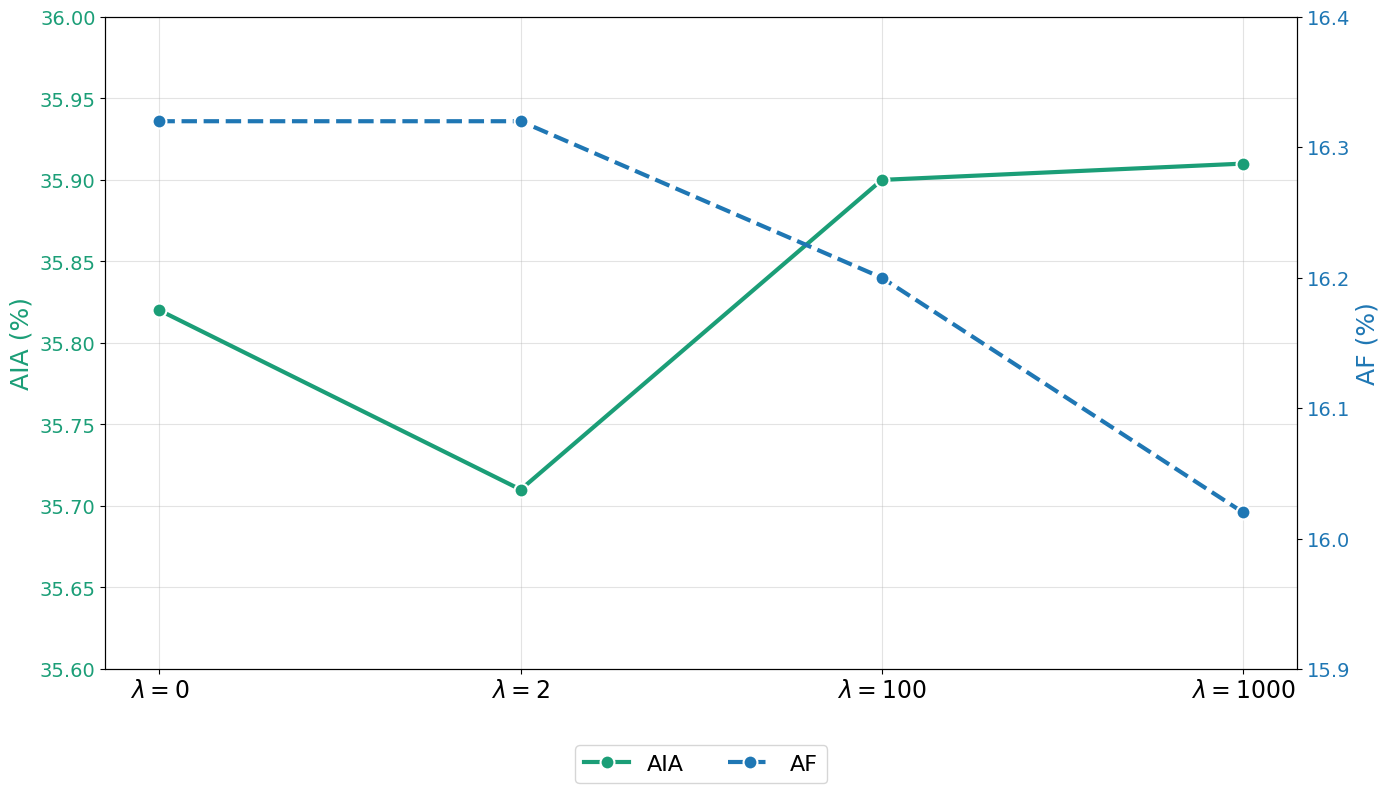}
    \caption{AIA and AF under different values of $\lambda$ on TinyImageNet in the CIL setting: $\lambda= 0$ (no neighbors),$\lambda = 2$, $\lambda = 100$, and $\lambda = 1000$.}
    \label{fig:lambda}
\end{figure}

\para{Ablation on Neighboring Regions} Since neighboring parameters represent one of the key components of SPARS, this experiment investigates their contribution to knowledge preservation under different regularization strengths. Specifically, we further explored the effect of the neighbors and the strength of regularization by investigating the following values of the regularization coefficient $\lambda $: [0, 2, 100, 1000]  under the CIL setting on TinyImageNet, with top-k set to 100k. \autoref{fig:lambda} presents the AIA and AF across different $\lambda $ values. AF decreases as $\lambda $ increases, indicating that stronger regularization leads to better preservation of previous knowledge. At the same time, AIA remains stable and slightly improves as $\lambda $ increases, suggesting that stronger regularization does not negatively affect plasticity. Although adding neighboring parameters results in limited changes in AIA, differences are observed in AF, indicating their contribution to knowledge retention. To better understand this behavior, the distributions of the top-k winner parameters and their neighboring parameters were further analyzed. 

\medskip

\begin{figure}[t]
    \centering
\includegraphics[width=0.80\linewidth]{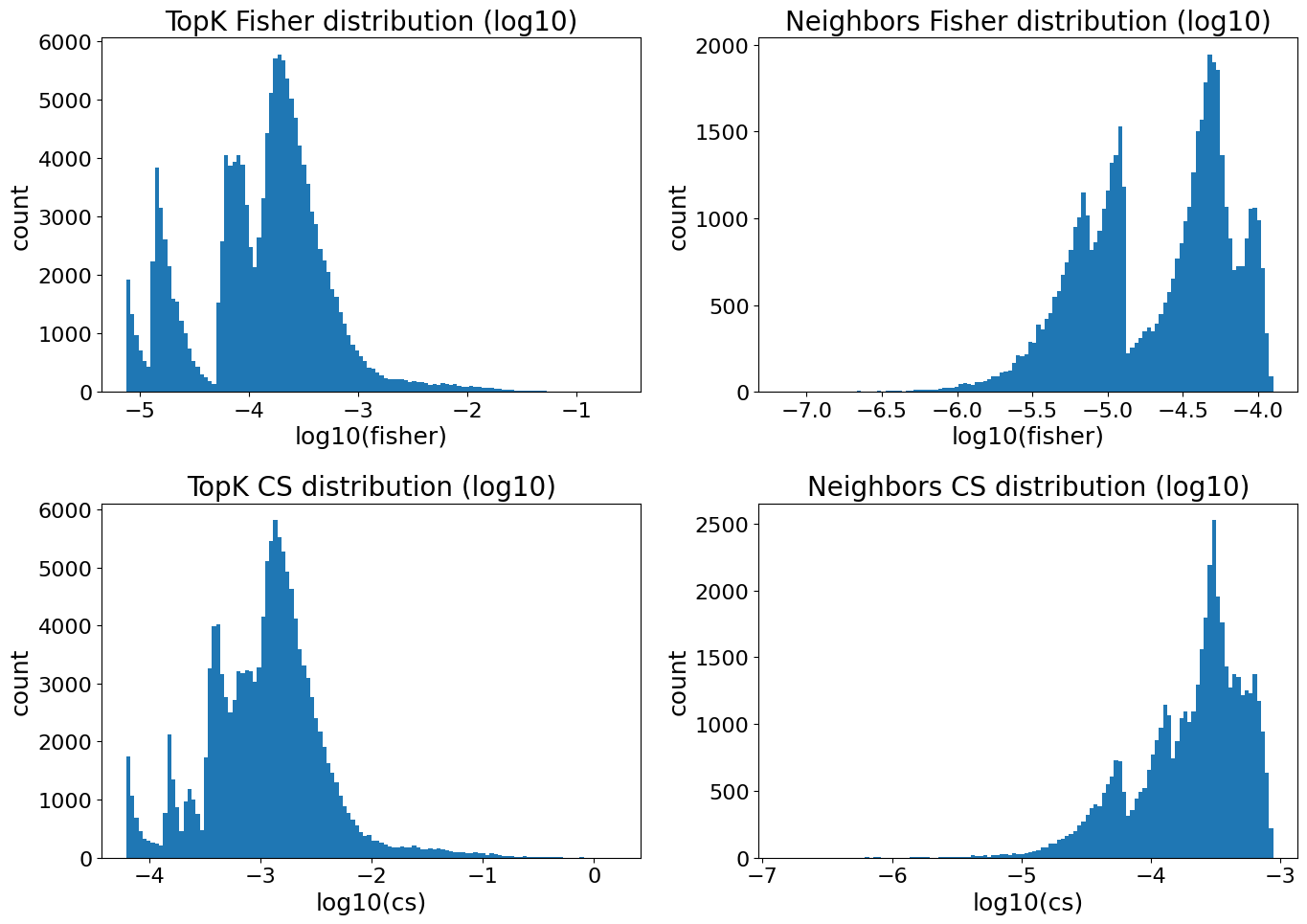}
    \caption{Distributions of FIM and CS (log10 scale) for the top-\(k\) winner parameters and their neighboring parameters on CIFAR-10 in the CIL setting.}
    \label{fig:Fisher Distribution}
\end{figure}

\para{Distribution of importance Scores} \autoref{fig:Fisher Distribution} shows that the importance values of the top-k parameters exhibit a highly skewed distribution with significantly larger importance values. Especially in the tail of the distribution, only a small number of parameters possess very high importance values, indicating that the learning process across all 10 tasks is sparse. On the other hand, the neighbors exhibit lower importance values and a denser distribution. Nevertheless, the two distributions show considerable overlap, suggesting that neighboring parameters are also important to some extent, which supports our hypothesis that neighbors contain meaningful information and contribute to knowledge. This means that even after removing the exact overlap between the neighbor list and the top-k list, the importance values of the remaining neighbors still remain close to many of the selected top-k parameters. This observation indicates that the importance estimation method can capture the importance of neighboring parameters and suggests that knowledge in CNNs is structurally distributed, where neighboring parameters share importance and collaboratively support learning. Consequently, learning in convolutional layers may not be entirely sparse at the kernel level, particularly when using relatively small convolutional kernels (e.g., 3×3). We further explored the ratio of the exact overlap and found that more than 80\% of the top-k list intersects with the neighbor set. This partially explains why the effect of neighboring parameters remains limited in AIA. However, it still affects AF, which suggests that neighbor regularization mainly improves stability by supporting the top-k parameters. In this context, the ratio between the regularization gradient and the cross-entropy gradient was further analyzed. We found that the regularization gradient is approximately 0.00138\% of the cross-entropy gradient, making it too weak to effectively dominate the optimization process in favor of preserving neighbors. This behavior is partially influenced by $\lambda$; therefore, increasing $\lambda$ improves AF. On the other hand, the small importance values of the neighbors also contribute to weak regularization strength.

\medskip

\begin{figure}[t]
    \centering
\includegraphics[width=0.80\linewidth]{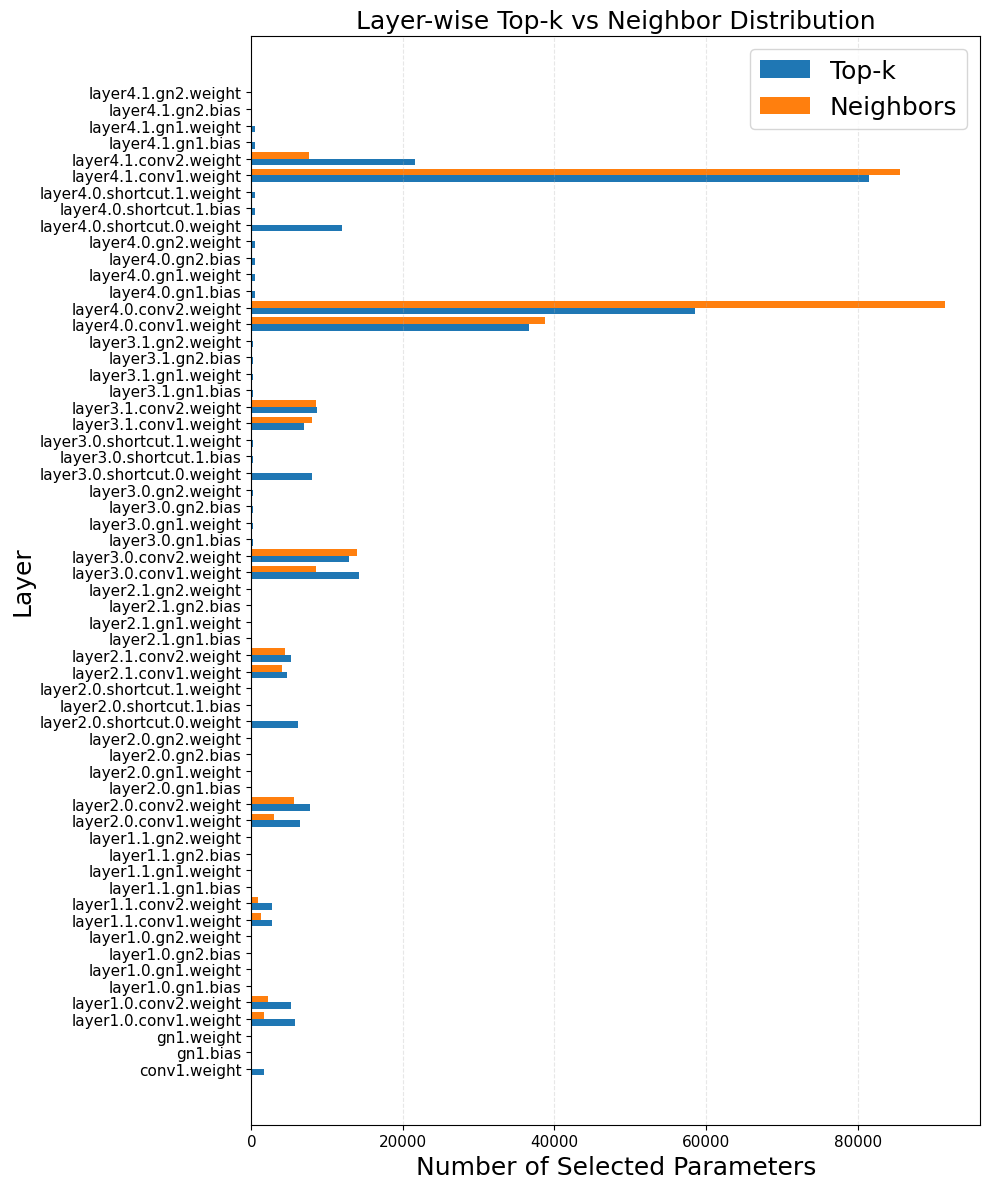}
    \caption{Layer-wise distribution of top-k winner and neighboring parameters across convolutional layers.}
    \label{fig:layers-distribution}
\end{figure}

\para{Layer-Wise Parameter Distribution} \autoref{fig:layers-distribution} shows that the number of neighbors in the early layers is smaller than the number of top-k parameters, which indicates a high exact overlap between them before overlap removal. In these layers, the kernels contain a higher proportion of top-k parameters compared to neighbors, suggesting that a large fraction of the kernel parameters contribute significantly to learning low-level image representations and therefore receive high Fisher importance scores. In contrast, the deeper layers contain a larger neighboring set even after removing the overlap, which may indicate that the overlap ratio is lower in these layers and that learning becomes more sparse. In other words, not all parameters contribute equally in the deeper layers. Since these layers are no longer responsible for capturing fine image details, they instead focus on learning more abstract representations.

Overall, these observations help explain why the impact of neighboring parameters is less visible. In addition, the relatively small kernel size (3×3) may further reduce the observable effect of neighboring parameters. The high overlap ratio between the winner and neighboring parameters further supports the importance of neighboring parameters and provides additional insight into the behavior of convolutional layers, which is consistent with our hypothesis. On the other hand, our decision to handle the overlap by removing it from the neighbor list may further reduce its impact in terms of AIA. In this context, we conducted exploratory experiments aimed at avoiding the overlap by performing local top-k selection, where any kernel containing a selected winner parameter is excluded from further top-k selection. As a result, the overlap ratio becomes zero, but at the expense of AIA, which slightly decreases because this strategy ignores the importance values during the selection process. Moreover, no remarkable improvements were observed in the ablation study.

\section{Conclusion}

This work introduces a hybrid continual learning framework that mitigates catastrophic forgetting while maintaining a balance between stability and plasticity without using a memory buffer, model expansion, or requiring training data from previous tasks.
The proposed SPARS framework bridges parameter isolation and parameter regularization by partitioning the network into three importance-aware regions in which each region is assigned a dedicated adaptation strategy. 
This region-aware adaptation attempts to obtain a better stability-plasticity trade-off than existing methods.
Experimental results demonstrate superior  performance on TinyImageNet under both task-incremental learning and class-incremental learning settings, as well as competitive performance on CIFAR-10. The findings also indicate that performance degradation in CIL is influenced not only by feature drift but also by classifier behavior, as evidenced by the superior performance of statistical classifiers. Moreover, the Confidence Score performs competitively with the Fisher Information Matrix, highlighting its effectiveness in privacy-sensitive and data-restricted scenarios. These findings demonstrate that importance-aware region selection provides an effective and scalable direction for memory-free continual learning in real-world applications.

\medskip

\para{Limitations}

The framework heavily relies on the importance estimation strategy and method, and it also depends on a convolutional based architecture.
In this work, neighborhood relationships are investigated at the kernel parameter level using a fixed, predefined number of neighbors. Therefore, investigating different types of relationships, such as those at the filter or channel level, and exploring more flexible selection approaches may improve performance.
In addition, the regularization term regularizes the neighbors individually based on their importance scores.
Therefore, global regularization that preserves the selected neighborhood area and mitigates the overall shift may help preserve the relationships among neighboring parameters and further support the knowledge represented in these areas.
Furthermore, the substantial overlap between the neighbors and the top-\(k\) parameters may indicate that the selection of the top-\(k\) parameters should follow specific criteria, such as satisfying a predefined threshold, to avoid including too many parameters that may not contribute to stability.

\medskip

\appendix
\renewcommand{\thesection}{Appendix \Alph{section}}
\section{Baseline descriptions}
\label{app:baselines}
This appendix provides brief descriptions of the baseline methods considered in our evaluation. For clarity, the baselines are organized according to their primary continual learning strategy, including regularization-based, isolation-based, hybrid, memory replay-based, and pseudo-replay methods. For each baseline, we summarize its main mechanism for preserving previously acquired knowledge and mitigating catastrophic forgetting. 

\label{app:baselines}
\textbf{Regularization-Based Method:}
\begin{itemize}
    \item oEWC~\cite{schwarz2018progress}, Online Elastic Weight Consolidation, estimates parameter importance using the FIM in an online manner and regularizes important parameters against changes.
    \item SI~\cite{zenke2017continual}, Synaptic Intelligence, estimates parameter importance by monitoring the influence of each parameter on the loss function throughout the training phase in an online manner, and accumulate these importance scores along the learning path to obtain a more robust estimate of parameter importance. The changes to important parameters are then penalized.
    \item MAS~\cite{aljundi2018memory}, Memory Aware Synapses, estimates the importance of model parameters in an unsupervised manner by measuring the sensitivity of the predicted output function to changes in each parameter. Changes to important parameters are then penalized.
\end{itemize}

\textbf{Isolation-Based Methods:}

\begin{itemize}

    \item SupSup~\cite{Wortsman2020Supermasks}, Supermasks in Superposition, learns a task-specific subnetwork on top of a fixed, randomly initialized network. When the task identity is not available, the framework formulates task inference as an optimization problem, where the correct subnetwork is expected to produce a low-entropy output distribution.
    \item 
    WSN~\cite{kang2023forget}, Winning SubNetworks, Finds subnetworks based on binary masks by selecting the top-p parameters that are most important for each task to construct the corresponding subnetwork.
    Parameter importance is determined by a learnable importance weight associated with each regular model weight.
    \item SPG~\cite{konishi2023parameter}, Soft-masking of Parameter-level Gradient flow, Creates subnetworks in a soft manner, where parameter importance is determined based on gradients.
    The gradient flow of these parameters is directly constrained during the backward pass, restricting updates to each parameter individually according to its importance score.
    \item
    IBM~\cite{chen2023towards}, Information Bottleneck Masked Sub-network, accumulates important information into weights based on the information bottleneck principle to create redundancy-free subnetworks.
    It mitigates catastrophic forgetting by freezing these weights via task-specific binary masks.
\end{itemize}

\textbf{Hybrid Methods:}

\begin{itemize}

    \item
    EDD~\cite{moon2026expandable}, Expandable and Differentiable Dual Memory, consists of two memories: one learns common features that can be shared across all tasks, while the other is task-specific and learns discriminative features.
    Orthogonal regularization is applied to the task-specific memory.
    \item
    HAL~\cite{chaudhry2020using}, Hindsight Anchor Learning, Uses stored samples to approximate forgetting through fine-tuning and optimize the anchor points computed for each task.
    These anchor points are then preserved through regularization.
\end{itemize}

\textbf{Memory Replay-Based Methods:}

\begin{itemize}
  
    \item
    iCaRL~\cite{rebuffi2017icarl}, Incremental Classifier and Representation Learning, utilizes KD and prototype rehearsal to preserve previously learned knowledge and classifies samples using the nearest-mean-of-exemplars rule.
    \item
    DER~\cite{Buzzega2020DarkExperienceReplay}, Dark Experience Replay, Stores past samples along with their corresponding logits produced during training.
    Knowledge is preserved by aligning the current model outputs with the stored logits using an $L^2$ matching loss. 
    \item
    BiC~\cite{wu2019bic}, Bias Correction, addresses the bias issue in CIL by training a correction layer with two learnable parameters on a validation set to reduce the bias toward new classes.
    \item
     WA~\cite{zhao2019maintaining}, Weight Aligning, Uses stored samples and KD to preserve the discrimination of old classes, and then aligns the classifier weights as a post-processing step after training.
    \item
    TPCIL~\cite{Tao2020TPCIL}, Topology-Preserving Class-Incremental Learning, Uses an Elastic Hebbian Graph to model the topology of the feature space and proposes a topology-preserving loss to penalize changes in feature relationships.

    \item
    
    MER~\cite{riemer2018learning}, Meta-Experience Replay, Combines experience replay with optimization-based meta-learning.
    It uses the Reptile algorithm in the inner and outer loops to update the parameters. The inner loop maintains alignment within each batch, while the outer loop maintains alignment across batches, ensuring that the updates do not interfere with previous tasks while maximizing the transfer of previously learned knowledge.

    \item 
    GEM~\cite{lopezpaz2017gradient}, Gradient Episodic Memory, Uses an episodic memory of previous tasks and computes the gradients of both the current and previous tasks.
    It uses the gradients from previous tasks to define constraints on the current update, ensuring that it does not increase their losses.
    When the constraints are violated, GEM modifies the current gradient by solving a quadratic programming problem.
    \item
    a-GEM~\cite{chaudhry2019efficient}, Averaged GEM, simplifies GEM by replacing its multiple task-specific constraints with a single constraint based on the average loss over the episodic memory, significantly reducing computational and memory costs.
    \item
    UER~\cite{wang2024uncertainty}, Uncertainty-Aware Dark Experience Replay, improves upon DER by leveraging model and data uncertainty to select representative samples through an Uncertainty-Aware Sampling strategy. To estimate uncertainty, it employs a Bayesian Neural Network.
    In addition, it introduces an uncertainty loss into DER to preserve the uncertainty information associated with previous tasks, thereby further mitigating catastrophic forgetting.
\end{itemize}

\textbf{Pseudo-Replay Methods:}

\begin{itemize}
    \item
    DGM\cite{Ostapenko2019LearningToRemember}, Dynamic Generative Memory, employs a generative memory module that synthesizes samples from previous classes. The method learns task-specific binary masks to control plasticity in the generator. Additionally, DGM dynamically expands the network capacity to maintain a consistent number of free parameters for each task.
    \item
    DGR~\cite{Shin2017DeepGenerativeReplay}, Deep Generative Replay, employs a generative replay strategy in which a ”scholar” model, composed of a generator and a solver, is jointly trained.
    The generator synthesizes samples of previously learned tasks, while the solver learns from a mixture of real data from the current task and replayed synthetic data.
    At each new task, the generator reconstructs the cumulative input distribution, and the solver aligns its predictions with both real labels and past solver outputs. 
    \item
     MeRGANs~\cite{Wu2018MemoryReplayGANs}, Memory Replay GANs, Uses a conditional GAN to generate samples from previous tasks and mitigates forgetting through two main mechanisms: joint training with replay and replay alignment.
\end{itemize}

\medskip

\para{Funding}
No funding was received for conducting this study.

\para{Data availability}
The data that support the findings of this study are openly available through the cited sources and our code repository.
\section*{Declarations}

\noindent\textbf{Conflict of interest}\hspace{0.5em}
The authors declare no competing interests that are relevant to the content of this article.
\bibliography{sn-bibliography}

\end{document}